\documentclass[11pt]{article}
\usepackage{eacl2017}
\usepackage{times}
\usepackage{url}
\usepackage{latexsym}
\usepackage{times}
\usepackage{url}
\usepackage{latexsym}
\usepackage{graphicx}
\usepackage{multirow}
\usepackage{booktabs}
\usepackage{pgfplots}
\usepackage{lscape}
\usepackage{tikz}
\usetikzlibrary{arrows,positioning,shapes.geometric}
\usepackage{amsmath}
\usepackage{algorithm}
\usepackage[noend]{algpseudocode}
\usepackage{array,multirow}
\usepackage{algpseudocode}
\usepackage{pifont}
\usepackage{graphicx}

\eaclfinalcopy 
\def\eaclpaperid{20} 

\title{Word Similarity Datasets for Indian Languages: \\Annotation and Baseline Systems}

\author{Syed S. Akhtar\qquad Arihant Gupta\qquad Avijit Vajpayee \qquad Arjit Srivastava\qquad M. Shrivastava\\International Institute of Information Technology\\ Hyderabad, Telangana, India\\ \{syed.akhtar, arihant.gupta, arjit.srivastava\}@research.iiit.ac.in,\\ avijit.vajpayee@students.iiit.ac.in,\\ manish.shrivastava@iiit.ac.in}

\date{}

\begin{document}
\maketitle
\begin{abstract}

With the advent of word representations, word similarity tasks are becoming increasing popular as an evaluation metric for the quality of the representations. In this paper, we present manually annotated monolingual word similarity datasets of six Indian languages - Urdu, Telugu, Marathi, Punjabi, Tamil and Gujarati. These languages are most spoken Indian languages worldwide after Hindi and Bengali. For the construction of these datasets, our approach relies on translation and re-annotation of word similarity datasets of English. We also present baseline scores for word representation models using state-of-the-art techniques for Urdu, Telugu and Marathi by evaluating them on newly created word similarity datasets.

\end{abstract}

\section{Introduction}

Word representations are being increasingly popular in various areas of natural language processing like dependency parsing ~\cite{BSL:14}, named entity recognition ~\cite{MIL:04} and parsing ~\cite{SOCH:13}. Word similarity task is one of the most popular benchmark for the evaluation of word representations. Applications of word similarity range from Word Sense Disambiguation~\cite{SP:05}, Machine Translation Evaluation~\cite{AL:09}, Question Answering~\cite{MM:11}, and Lexical Substitution~\cite{DM:09}.

Word Similarity task is a computationally efficient method to evaluate the quality of word vectors. It relies on finding correlation between human assigned semantic similarity (between words) and corresponding word vectors. We have used Spearman's Rho for calculating correlation. Unfortunately, most of the word similarity tasks have been majorly limited to English language because of availability of  well annotated different word similarity test datasets and large corpora for learning good word representations, where as for Indian languages like Marathi, Punjabi, Telugu etc - which even though are widely spoken by significant number of people, are still computationally resource poor languages. Even if there are models trained for these languages, word similarity datasets to test reliability of corresponding learned word representations do not exist.

Hence, primary motivation for creation of these six word similarity datasets has been to provide necessary evaluation resources for all the current and future work in field of word representations on these six Indian languages - all ranked in top 25 most spoken languages in the world, since no prior word similarity datasets have been publicly made available. 

The main contribution of this paper is the set of newly created word similarity datasets which would allow for fast and efficient comparison between. Word similarity is one of the most important evaluation metric for word representations and hence as an evaluation metric, these datasets would promote development of better techniques that employ word representations for these languages. We also present baseline scores using state-of-the-art techniques which were evaluated using these datasets.

The paper is structured as follows. We first discuss the corpus and techniques used for training our models in section 2 which are later used for evaluation. We then talk about relevant related work that has been done with respect to word similarity datasets in Section 3. We then move on to explain how these datasets have been created in section 4 followed by our evaluation criteria and experimental results of various models evaluated on these datasets in Section 5. Finally, we analyze and explain the results in section 6 and finish this paper with how we plan to extend our work in section 7.  

\section{Datasets}
For all the models trained in this paper, we have used the Skip-gram, CBOW ~\cite{MSG:13a} and  FastText~\cite{PJ:16} algorithms. The dimensionality has been fixed at 300 with a minimum count of 5 along with negative sampling. 

As training set of Marathi, we use the monolingual corpus created by IIT-Bombay. This data contains 27 million tokens. For Urdu, we use the untagged corpus released by Jawaid et. al. ~\shortcite{BJ:15} containing 95 million tokens. For Telugu, we use Telugu wikidump available at ``https://archive.org/details/tewiki-20150305" having 11 million tokens. 

For testing, we use the newly created datasets. The word similarity datatsets for Urdu, Marathi, Telugu, Punjabi, Gujarati and Tamil contain 100, 104, 111, 143, 163 and 97 word pairs respectively. 

For rest of the paper, we have calculated the Spearman $\rho$ (multiplied by 100) between human assigned similarity and cosine similarity of our word embeddings for the word-pairs. For any word which was is not found, we assign it a zero vector.


In order to learn initial representations of the words, we train word embeddings (word2vec) using the parameters described above on the training set.

\section{Related Work}

Multitude of word similarity datasets have been created for English, like WordSim-353~\cite{FT:02}, MC-30~\cite{MC:91}, Simlex-999~\cite{FH:16}, RG-65  ~\cite{RG:65} etc. RG-65 is one of the oldest and most popular datasets, being used as a standard benchmark for measuring reliability of word representations. 

RG-65 has also acted as base for various other word similarity datasets created in different languages : French ~\cite{JI:11}, German~\cite{ZG:06},Portuguese~\cite{RG:14}, Spanish and Farsi ~\cite{CC:15}. While German and Portuguese reported IAA (Inter Annotator Agreement) of 0.81 and 0.71 respectively, no IAA was calculated for French. For Spanish and Farsi, inter annotator agreement of 0.83 and 0.88 respectively was reported. Our datasets were created using RG-65 and WordSim-353 as base, and their respective IAA(s) are mentioned later in the paper.  

\section{Construction of Monolingual Word Similarity datasets}

\subsection{Translation}

English RG-65 and WordSim-353 were used as base for creating all of our six different word similarity datasets. Translation of English data set to target language (one of the six languages) was manually done by a set of three annotators who are native speakers of the target language and are fluent in English. Initially, translations are provided by two of them, and in case of disparity, third annotator was used as a tie breaker.

Finally, all three annotators reached a final set of translated word pairs in target language, ensuring that there were no repeated word pairs. This approach was followed by Camacho-Callados et al. \shortcite{CC:15} where they created word similarity datasets for Spanish and Farsi in a similar manner.

\subsection{Scoring}

For each of the six languages, 8 native speakers were asked to manually evaluate each word similarity data set individually. They were instructed to indicate, for each pair, their opinion of how similar in meaning the two words are on a scale of 0-10, with 10 for words that mean the same thing, and 0 for words that mean completely different things. The guidelines provided to the annotators were based on the SemEval task on Cross-Level Semantic Similarity~\cite{DJ:14}, which provides clear indications in order to distinguish similarity and relatedness.

The results were averaged over the 8 responses for each word similarity data set, and each data set saw good agreement amongst the evaluators, except for Tamil, which saw relatively weaker agreement with respect to other languages (see Table~\ref{IAA Score}). 
\section{Evaluation}

\subsection{Inter Annotator Agreement (IAA)}
The meaning of a sentence and its words can be interpreted in different ways by different readers. This subjectivity can also reflect in annotation of sentences of a language despite the annotation guidelines being well defined. Therefore, inter-annotator agreement is calculated to give a measure of how well the annotators can make the same annotation decision for a certain category. 
\begin{table}[h]
\begin{center}
\begin{tabular}{|c|c|}
\hline \bf Language &\bf Inter Annotator Agreement \\ \hline
Urdu &  0.887  \\
Punjabi & 0.821 \\
Marathi & 0.808  \\
Tamil & 0.756 \\
Telugu & 0.866 \\
Gujarati & 0.867 \\
\hline
\end{tabular}
\end{center}
\caption{\label{IAA Score}Inter Annotator Agreement (Fleiss Kappa) scores for word similarity datasets created for six languages. }
\end{table}

\subsubsection{Fleiss' Kappa}
Fleiss' kappa is a statistical measure for assessing the reliability of agreement between a fixed number of raters when assigning categorical ratings to a number of items or classifying items. This contrasts with other kappas such as Cohen's kappa, which only work when assessing the agreement between not more than two raters or the interrater reliability for one appraiser versus himself. The measure calculates the degree of agreement in classification over that which would be expected by chance \cite{Wiki:17}.

We have calculated Fleiss' Kappa for all our word similarity datasets (see table~\ref{IAA Score}).

\section{Result and Analysis}

\begin{table}[h]
\begin{center}
\begin{tabular}{|c|c|c|c|}
\hline \bf System &\bf  Score &\bf OOV &\bf Vocab \\ \hline
CBOW &  28.30  & 19 & 130K \\
SG & 34.40 & 19 & 130K \\
FastText & 34.61 & 19 & 130K  \\
FastText w/ OOV & 45.47 & 14 & - \\
\hline
\end{tabular}
\end{center}
\caption{\label{Urdu Results}Results for \textbf{Urdu} }
\end{table}

\begin{table}[h]
\begin{center}
\begin{tabular}{|c|c|c|c|}
\hline \bf System &\bf  Score &\bf OOV &\bf Vocab \\ \hline
CBOW &  36.16  & 3 & 194K \\
SG & 41.22 & 3 & 194K \\
FastText & 33.68 & 3 & 194K  \\
FastText w/ OOV & 38.66 & 0 & - \\
\hline
\end{tabular}
\end{center}
\caption{\label{Marathi Results}Results for \textbf{Marathi} }
\end{table}

\begin{table}[h]
\begin{center}
\begin{tabular}{|c|c|c|c|}
\hline \bf System &\bf  Score &\bf OOV &\bf Vocab \\ \hline
CBOW &  26.01  & 14 & 174K \\
SG & 27.04 & 14 & 174K \\
FastText & 34.29 & 14 & 174K  \\
FastText w/ OOV & 46.02 & 0 & - \\
\hline
\end{tabular}
\end{center}
\caption{\label{Telugu Results}Results for \textbf{Telugu} }
\end{table}

We present baseline scores using state of the art techniques - CBOW and Skipgram ~\cite{MSG:13a} and FastText-SG ~\cite{PJ:16}, evaluated using our word similarity datasets in tables~\ref{Urdu Results},~\ref{Marathi Results} and ~\ref{Telugu Results}. As we can see the models trained encountered unseen word pairs when evaluated on their corresponding word similarity datasets. This goes on to show that all word pairs in our word similarity sets are not too common, and contain word pairs with some rarity.

We see that FastText w/ OOV (Out of Vocabulary) performed better than FastText in all the experiments, because character based models perform better than rest of the models since they are able to handle unseen words by generating word embeddings for missing words via character model.
\section{Future Work}

There are a lot of Indian languages that are still  computationally resource poor even though they are widely spoken by significant number of people. Our work is a small step towards generating resources to further the research involving word representations on Indian languages.

To further extend our work, we will create rare-word word similarity datasets for six languages we worked on in this paper, and creating word similarity datasets for other major Indian languages as well.

We will also work on improving word representations for the languages we worked on, hence improve the baseline scores that we present here. This will require us to build new corpus to train our models for three languages that we couldn't provide baseline scores for - Punjabi, Tamil and Gujarati and build more corpus for Urdu, Telugu and Marathi to train better word embeddings.

\end{document}